\documentclass[lettersize,journal,twoside]{IEEEtran}
\usepackage{amsmath,amsfonts,amssymb}
\usepackage{algorithm}
\usepackage{algpseudocode}
\usepackage{array}
\usepackage[caption=false,font=normalsize,labelfont=sf,textfont=sf]{subfig}
\usepackage{textcomp}
\usepackage{stfloats}
\usepackage{url}
\usepackage{verbatim}
\usepackage{graphicx}
\usepackage{cite}
\usepackage[export]{adjustbox}
\usepackage[table,xcdraw]{xcolor}
\usepackage{color}
\usepackage[hidelinks]{hyperref}
\definecolor{linkcolor}{HTML}{ee0000}
\hypersetup{
     colorlinks=true,
	 linkcolor=linkcolor,
	 filecolor=linkcolor,
	 urlcolor=linkcolor,
	 citecolor=linkcolor
}
\usepackage{multirow}
\usepackage{booktabs}
\usepackage{orcidlink}
\usepackage{threeparttable}
\usepackage{pifont}

\newcommand\liehat[1]{\left[ #1 \right]_\times}

\newcommand\mlcomment[1]{\iffalse #1 \fi}
\newcommand\bsm[1]{\boldsymbol{\mathrm{#1}}}
\newcommand\transform[2]{{\bsm{T}_{#1}^{#2}}}
\newcommand\transformhat[2]{{\hat{\bsm{T}}_{#1}^{#2}}}

\newcommand\rotation[2]{{\bsm{R}_{#1}^{#2}}}
\newcommand\rotationhat[2]{{\hat{\bsm{R}}_{#1}^{#2}}}

\newcommand\timeoffset[2]{{\tau_{#1}^{#2}}}
\newcommand\timeoffsethat[2]{{\hat{\tau}_{#1}^{#2}}}

\newcommand\angvel[2]{{\bsm{\omega}_{#1}^{#2}}}
\newcommand\angvelhat[2]{{\hat{\bsm{\omega}}_{#1}^{#2}}}

\newcommand\translation[2]{{\bsm{p}_{#1}^{#2}}}
\newcommand\translationhat[2]{{\hat{\bsm{p}}_{#1}^{#2}}}

\newcommand\linvel[2]{{\bsm{v}_{#1}^{#2}}}
\newcommand\linvelhat[2]{{\hat{\bsm{v}}_{#1}^{#2}}}

\newcommand\linacce[2]{{\bsm{a}_{#1}^{#2}}}
\newcommand\linaccehat[2]{{\hat{\bsm{a}}_{#1}^{#2}}}

\newcommand\gravity[1]{{\bsm{g}^{#1}}}
\newcommand\gravityhat[1]{{\hat{\bsm{g}}^{#1}}}

\newcommand\smallminus{{\text{-}}}
\newcommand\smallplus{{\text{+}}}
\newcommand\coordframe[1]{\underrightarrow{\mathcal{F}}_{#1}}
\newcommand\Exp[1]{\mathrm{Exp}\left( #1\right) }
\newcommand\Log[1]{\mathrm{Log}\left( #1\right) }

\newcommand{\figtabbottomspace}{\vspace{-15pt}}
\usepackage{siunitx}
\begin{document}

\title{eKalibr-Inertial: Continuous-Time Spatiotemporal Calibration for  Event-Based Visual-Inertial Systems}
\author{
Shuolong Chen \hspace{-1mm}$^{\orcidlink{0000-0002-5283-9057}}$
, Xingxing Li \hspace{-1mm}$^{\orcidlink{0000-0002-6351-9702}}$, 
 and Liu Yuan \hspace{-1mm}$^{\orcidlink{0009-0003-6039-7070}}$ 




\thanks{The authors are with the School of Geodesy and Geomatics (SGG), Wuhan University (WHU), Wuhan 430070, China.
Corresponding author: Xingxing Li (\texttt{xxli@sgg.whu.edu.cn}). 
The specific contributions of the authors to this work are listed in Section \hyperref[sect:author_contribution]{\textbf{CRediT Authorship Contribution Statement}} at the end of the article.}	

}
\markboth{IEEE Robotics and Automation Letters. Preprint Version. Accepted
August, 2025}
{Chen \MakeLowercase{\textit{et al.}}: eKalibr-Stereo: Continuous-Time Spatiotemporal Calibration for Event-Based Stereo Visual Systems}


\maketitle

\begin{abstract}
The bioinspired event camera, distinguished by its exceptional temporal resolution, high dynamic range, and low power consumption, has been extensively studied in recent years for motion estimation, robotic perception, and object detection.
In ego-motion estimation, the visual-inertial setup is commonly adopted due to complementary characteristics between sensors (e.g., scale perception and low drift).
For optimal event-based visual-inertial fusion, accurate spatiotemporal (extrinsic and temporal) calibration is required.
In this work, we present \emph{eKalibr-Inertial}, an accurate spatiotemporal calibrator for event-based visual-inertial systems, utilizing the widely used circle grid board.
Building upon the grid pattern recognition and tracking methods in \emph{eKalibr} and \emph{eKalibr-Stereo},
the proposed method starts with a rigorous and efficient initialization, where all parameters in the estimator would be accurately recovered.
Subsequently, a continuous-time-based batch optimization is conducted to refine the initialized parameters toward better states.
The results of extensive real-world experiments show that \emph{eKalibr-Inertial} can achieve accurate event-based visual-inertial spatiotemporal calibration.
The implementation of \emph{eKalibr-Inertial} is open-sourced at (\url{https://github.com/Unsigned-Long/eKalibr}) to benefit the research community.
\end{abstract}

\begin{IEEEkeywords}
Event camera, inertial measurement unit, spatiotemporal calibration, continuous-time optimization
\end{IEEEkeywords}

\section{Introduction and Related Works}
\IEEEPARstart{B}{ioinspired} event cameras have attracted considerable research interest in recent years, due to their advantages of low sensing latency and high dynamic range over conventional standard (frame-based) cameras \cite{guan2023pl}.
The ego-motion estimation in high-dynamic-range and high-speed scenarios is one of applications of the event camera, where a visual-inertial setup is commonly employed \cite{chen2023esvio,wang2025asyneio,LU-RSS-24}.
For such an event-based stereo visual sensor suite, accurate spatiotemporal calibration is required to determine extrinsics and time offset between cameras for subsequent data fusion.

Visual-inertial spatiotemporal calibration typically consists of two sub-modules: ($i$) correspondence construction (front end) and ($ii$) spatiotemporal optimization (back end).
In the front end, artificial visual targets, such as checkerboards \cite{yu2006robust}, April Tags \cite{wang2016apriltag},  and ChArUco board \cite{an2018charuco}, are commonly employed to construct accurate 3D-2D correspondences with real-world geometric scale through pattern recognition.
While a substantial number of target pattern recognition methods \cite{yu2006robust,sun2008robust,hu2019deep} oriented to standard cameras have been proposed, they are not applicable to event cameras, which output
asynchronous event stream rather than conventional intensity images.
To recognize target patterns from raw events, early works \cite{rpg_dvs_ros_calibration,dominguez2019bio,cai2024accurate} generally rely on blinking light emitting diode (LED) grid boards.
Although target patterns can be accurately extracted, requiring additional LED boards introduces inconvenience.
Meanwhile, these methods typically require the event camera to remain stationary, making them unsuitable for visual-inertial spatiotemporal calibration that necessitates motion excitation \cite{chen2025ikalibr}.
To address this, subsequent methods \cite{muglikar2021calibrate,jiao2023lce} have proposed an alternative approach, namely reconstructing intensity images from raw events using event-based image reconstruction methods (such as E2VID\cite{rebecq2019high} and Spade-E2VID \cite{cadena2021spade}) first, followed by conventional image-based pattern recognition methods.
Although reconstructed images exhibit high consistency, substantial noise within the images could lead to imprecise pattern extraction, which further affects calibration accuracy.
Considering these, some event-based pattern recognition methods have been proposed recently, aiming to extract target patterns from dynamically acquired raw events directly.
A typical work is our previously proposed \emph{eKalibr} \cite{chen2025ekalibr} (event camera intrinsic calibration), which clusters events and matches clusters based on normal flow estimation, enabling efficient and accurate event-based circle grid pattern extraction.
The follow-up stereo spatiotemporal calibrator \emph{eKalibr-Stereo} \cite{chen2025ekalibr-stereo} extends \emph{eKalibr} by incorporating a tracking module for incomplete grid patterns, thereby enhancing the continuity of target extraction.
As a subsequent effort, the present work inherits the event-based circle grid pattern recognition and tracking methods developed in our two earlier works.

In terms of the back end of the visual-inertial calibration, namely spatiotemporal optimization, event-based and frame-based calibrations share the same algorithmic framework, aiming to estimate spatiotemporal parameters using extracted visual target patterns and inertial measurements.
In general, spatiotemporal optimization can be categorized into discrete-time-based and continuous-time-based ones.
Discrete-time-based methods represent states using discrete estimates that are temporally coupled to measurements.
Based on the extended Kalman filter (EKF), Mirzaei et al. \cite{mirzaei2008kalman} proposed a visual-inertial extrinsic calibration method to determine the transformation between a standard camera and an inertial measurement unit (IMU).
Similarly, Hartzer et al. \cite{hartzer2022online} presented an EKF-based online visual-inertial extrinsic calibration method.
Yang et al. \cite{yang2016monocular} designed a sliding-window-based visual-inertial state estimator, supporting online camera-IMU extrinsic calibration.
Different from the discrete-time-based methods, continuous-time-based ones represent time-varying states using time-continuous functions (such as B-splines), enabling state querying at any time instance, and thus are more suitable for temporal calibration.
The well-known \emph{Kalibr} \cite{furgale2013unified} proposed by Furgale et al. is the first continuous-time-based calibration framework, which employs B-splines for state representation and supports both extrinsic and temporal calibration for visual-inertial, multi-IMU, and multi-camera sensor suites.
\emph{Kalibr} is then extended by Huai et al. \cite{huai2022continuous} to support the rolling shutter cameras for readout time calibration.
In addition to vision-related calibration, the continuous-time state representation has also been widely employed in other multi-sensor calibration, such as LiDAR-IMU \cite{lv2022observability} and radar-IMU \cite{chen2024ris} calibration.

In this article, focusing on event-based visual-inertial systems, we present a continuous-time-based spatiotemporal calibration method, named \emph{eKalibr-Inertial}, to accurately estimate the extrinsics and time offset between the event camera and IMU.
Building upon \emph{eKalibr} \cite{chen2025ekalibr} and \emph{eKalibr-Stereo} \cite{chen2025ekalibr-stereo}, \emph{eKalibr-Inertial} tracks continuous circle grid patterns (complete and incomplete ones) from raw events for 3D-2D correspondence construction.
Given the high non-linearity of continuous-time optimization, a three-stage initialization procedure is first conducted to recover the initials of states, which are then iteratively refined to optimal ones using a continuous-time-based batch bundle adjustment.
\emph{eKalibr-Inertial} makes the following (potential) contributions:
\begin{enumerate}
\item We proposed a continuous-time-based spatial and temporal calibrator for event-based visual-inertial systems, which could accurately determine both extrinsics and time offset of a event-based visual-inertial system.
To the best of our knowledge, this is the first open-source work focused on event-based visual-inertial spatiotemporal calibration.

\item Sufficient real-world experiments were conducted to comprehensively evaluate the proposed \emph{eKalibr-Inertial}.
Both the dataset and code implementation are open-sourced, to benefit the robotic community if possible.
\end{enumerate}

Note that the proposed \emph{eKalibr-Inertial} supports \textbf{one-shot} event-based \textbf{multi-camera multi-IMU} spatiotemporal calibration (an arbitrary number of event cameras and IMUs).
To enhance clarity, this article only considers the minimal configuration, i.e., sensor suite with an event camera and an IMU, as it's the most typical sensor setup for facilitating multi-camera multi-IMU calibration.

\section{Preliminaries}
This section presents notations and definitions utilized in
this article. The involved sensor intrinsic models (for the camera and IMU) and B-spline-based time-varying state representation are also introduced for
a self-contained exposition of this work.

\subsection{Notations and Definitions}
Given a raw event $\bsm{e}$ generated by the event camera, we use $\tau\in\mathbb{R}$, $\bsm{x}\in\mathbb{Z}^2$, and $p\in\{\smallminus 1, \smallplus 1\}$ to represent its timestamp, pixel position, and polarity, respectively, i.e., $\bsm{e}\triangleq\{\tau,\bsm{x},p\}$.
The camera frame, IMU frame (body frame), and world frame (defined by the circle grid board) are represented as $\coordframe{c}$, $\coordframe{b}$, and $\coordframe{w}$, respectively.
The transformation from $\coordframe{b}$ to $\coordframe{w}$ are parameterized as the Euclidean matrix $\transform{b}{w}\in\mathrm{SE(3)}$, which is defined as:
\begin{equation}
\transform{b}{w}\triangleq\begin{bmatrix}
\rotation{b}{w}&\translation{b}{w}\\
\bsm{0}_{1\times 3}&1
\end{bmatrix}
\end{equation}
where $\rotation{b}{w}\in\mathrm{SO(3)}$ and $\translation{b}{w}\in\mathbb{R}^3$ are the rotation matrix and translation vector, respectively.
In terms of their high-order kinematics, we use $\angvel{b}{w}\in\mathbb{R}^3$, $\linvel{b}{w}\in\mathbb{R}^3$, and $\linacce{b}{w}\in\mathbb{R}^3$ to express the angular velocity, linear velocity, and linear acceleration of $\coordframe{b}$ with respect to and parameterized in $\coordframe{w}$, respectively.
Finally, we use $\hat{(\cdot)}$ and $\tilde{(\cdot)}$ to represent the state estimates and noisy quantities, respectively.

\subsection{Sensor Intrinsic Models}
The camera intrinsic model characterizes the visual projection process whereby 3D points in the camera coordinate frame are geometrically mapped onto the image plane to derive corresponding 2D pixels.
Adhering to our previously proposed \emph{eKalibr} \cite{chen2025ekalibr},  the intrinsic camera model comprising the pinhole projection model \cite{kannala2006generic} and radial-tangential distortion model \cite{tang2017precision} are employed in this work, which can be expressed as:
\begin{equation}
\label{equ:proj_func}
\bsm{x}_p=\pi_c\left(\bsm{p}^c,\mathcal{X}_{\mathrm{intr}}^{c}\right)\triangleq
\bsm{K}\left(\mathcal{X}_{\mathrm{proj}}^{c}\right)\cdot 
\bsm{d}\left(\bsm{p}^c,\mathcal{X}_{\mathrm{dist}}^{c}\right)
\end{equation}
with
\begin{equation}
\begin{gathered}
\mathcal{X}_{\mathrm{intr}}^{c}\triangleq\mathcal{X}_{\mathrm{proj}}^{c}\cup\mathcal{X}_{\mathrm{dist}}^{c}\\
\mathcal{X}_{\mathrm{proj}}^{c}\triangleq\left\lbrace
f_x,f_y,c_x,c_y
\right\rbrace,\;
\mathcal{X}_{\mathrm{dist}}^{c}\triangleq\left\lbrace
k_1,k_2,p_1,p_2
\right\rbrace
\end{gathered}
\end{equation}
where $\bsm{d}:\mathbb{R}^{3}\mapsto\mathbb{R}^{3}$ represents the distortion function distorting normalized image coordinates using distortion parameters $\mathcal{X}_{\mathrm{dist}}^{c}$;
$\bsm{K}\in\mathbb{R}^{2\times 3}$ denotes the intrinsic matrix organized by projection parameters $\mathcal{X}_{\mathrm{proj}}^{c}$;
$\pi:\mathbb{R}^{3}\mapsto\mathbb{R}^2$ is the projection function projecting 3D point $\bsm{p}^c$ onto the image plane as 2D point $\bsm{x}_{p}$;
$\mathcal{X}_{\mathrm{intr}}^{c}$ represents the camera intrinsic parameters comprising $\mathcal{X}_{\mathrm{proj}}^{c}$ and $\mathcal{X}_{\mathrm{dist}}^{c}$, which can be pre-calibrated using \emph{eKalibr}.

As for the IMU intrinsic model, taking into account the biases, scale factors, and nonorthogonality factors, we express it as:
\begin{equation}
\label{equ:imu_mearsing}
\begin{aligned}
\tilde{\bsm{a}}=
\pi_a\left(\bsm{a},\mathcal{X}_{\mathrm{intr}}^a \right)\;&\triangleq
\bsm{M}_a\cdot\bsm{a}+\bsm{b}_a+\bsm{\epsilon}_a
\\[-2pt]
\tilde{\bsm{\omega}}=
\pi_\omega\left( \bsm{\omega},\mathcal{X}_{\mathrm{intr}}^\omega \right) &\triangleq
\bsm{M}_\omega\cdot\bsm{\omega}+\bsm{b}_\omega+\bsm{\epsilon}_\omega
\end{aligned}
\end{equation}
with
\begin{equation}
\label{equ:imu_intri}
\begin{gathered}
\mathcal{X}_{\mathrm{intr}}^{b}\triangleq\mathcal{X}_{\mathrm{intr}}^a\cup\mathcal{X}_{\mathrm{intr}}^\omega\\
\mathcal{X}_{\mathrm{intr}}^a\triangleq\left\lbrace
\bsm{M}_a,\bsm{b}_a
\right\rbrace,\;
\mathcal{X}_{\mathrm{intr}}^\omega\triangleq\left\lbrace
\bsm{M}_\omega,\bsm{b}_\omega
\right\rbrace
\end{gathered}
\end{equation}
where $\bsm{a}$ and $\bsm{\omega}$ are ideal specific force and angular velocity, while $\tilde{\bsm{a}}$ and $\tilde{\bsm{\omega}}$ are noisy measurements;
$\bsm{M}_a$ and $\bsm{M}_\omega$ are upper triangular mapping matrices, introducing the scale factors $s_{(\cdot)}$ and non-orthogonality factors $\gamma_{(\cdot)}$:
\begin{equation}
\bsm{M}_a\triangleq\begin{bmatrix}
s_{a,1}&\gamma_{a,1}&\gamma_{a,2}\\
0&s_{a,2}&\gamma_{a,3}\\
0&0&s_{a,3}
\end{bmatrix}
,\;
\bsm{M}_\omega\triangleq\begin{bmatrix}
s_{\omega,1}&\gamma_{\omega,1}&\gamma_{\omega,2}\\
0&s_{\omega,2}&\gamma_{\omega,3}\\
0&0&s_{\omega,3}
\end{bmatrix}.
\end{equation}
$\bsm{b}_a$ and $\bsm{b}_\omega$ denote time-varying biases of the accelerometer and gyroscope respectively, and are considered as constants in the proposed \emph{eKalibr-Inertial} (as the collected data piece for calibration is short).
$\bsm{\epsilon}_a$ and $\bsm{\epsilon}_\omega$ are corresponding zero-mean Gaussian white noises of sensors.
The intrinsics of the accelerometer and gyroscope, i.e., $\mathcal{X}_{\mathrm{intr}}^a$ and $\mathcal{X}_{\mathrm{intr}}^\omega$, together constitute the IMU intrinsics $\mathcal{X}_{\mathrm{intr}}^b$, which would also be estimated in this work.

\subsection{Continuous-Time State Representation}
To efficiently fuse asynchronous data for multi-sensor spatiotemporal determination, especially for time offset calibration, the continuous-time state representation is employed in this work to represent the time-varying rotation and position of the IMU.
Compared with the conventional discrete-time representation generally maintaining discrete states at measurement times, the continuous-time representation models time-varying states using time-continuous functions, such as Gaussian process regression \cite{barfoot2014batch}, hierarchical wavelets \cite{anderson2014hierarchical}, and B-splines \cite{furgale2012continuous}, enabling state querying at arbitrary time.
In this work, the uniform B-spline is utilized for continuous-time state representation, which inherently possesses sparsity due to its local controllability, allowing computation acceleration in optimization \cite{chen2025ikalibr}.

The uniform B-spline is characterized by the spline order, a temporally uniformly distributed control point sequence, and a constant time distance between neighbor control points.
Specifically, given a series of translational control points:
\begin{equation}
\label{equ:pos_cp}
\begin{gathered}
\mathcal{X}_{\mathrm{pos}}\triangleq\left\lbrace
\bsm{p}_i,\tau_i\mid\bsm{p}_i\in\mathbb{R}^3,\tau_i\in\mathbb{R}
\right\rbrace 
\\
\mathrm{s.t.}\;\;
\tau_{i\smallplus 1}-\tau_i\equiv\Delta\tau_{\mathrm{pos}}
\end{gathered}
\end{equation}
the position $\bsm{p}(\tau)$ at time $\tau\in[\tau_i, \tau_{i\smallplus 1})$ of a $k$-order uniform B-spline can be computed as follows:
\begin{equation}
\label{equ:pos_bspline}
\begin{gathered}
\bsm{p}(\tau)=\bsm{p}_i+\sum_{j=1}^{k\smallplus 1}\lambda_j(u)\cdot\left(\bsm{p}_{i\smallplus j}-\bsm{p}_{i\smallplus j\smallminus 1} \right) 
\\
\mathrm{s.t.}\;\;
u=\frac{\tau-\tau_i}{\Delta\tau_{\mathrm{pos}}}
\end{gathered}
\end{equation}
where $\lambda_j(\cdot)$ denotes the $j$-th element of vector $\bsm{\lambda}(u)$ obtained from the order-determined cumulative
matrix and $u$ \cite{chen2025ikalibr}.
In this work, the cubic uniform B-spline ($k=4$) is employed.

The B-spline representation of time-varying rotation has similar forms with (\ref{equ:pos_bspline}) by replacing vector addition in $\mathbb{R}^3$ with group multiplication in $\mathrm{SO(3)}$.
The key distinction resides in the scalar multiplication operated within the Lie algebra $\mathfrak{so}(3)$, rather than on the Lie group manifold, to ensure closedness \cite{sommer2020efficient}.
Specifically, given a series of rotational control points:
\begin{equation}
\label{equ:rot_cp}
\begin{gathered}
\mathcal{X}_{\mathrm{rot}}\triangleq\left\lbrace
\bsm{R}_i,\tau_i\mid\bsm{R}_i\in\mathrm{SO(3)},\tau_i\in\mathbb{R}
\right\rbrace 
\\
\mathrm{s.t.}\;
\tau_{i\smallplus 1}-\tau_i\equiv\Delta\tau_{\mathrm{rot}}
\end{gathered}
\end{equation}
the rotation $\bsm{R}(\tau)$ at time $\tau\in[\tau_i, \tau_{i\smallplus 1})$ of a $k$-order uniform B-spline can be computed as follows:
\begin{equation}
\label{equ:rot_bspline}
\begin{gathered}
\bsm{R}(\tau)=\bsm{R}_i\cdot\prod_{j=1}^{k\smallplus 1}\mathrm{Exp}\left( \lambda_j(u)\cdot\mathrm{Log}\left(\bsm{R}_{i\smallplus j\smallminus 1}^\top\cdot\bsm{R}_{i\smallplus j} \right)\right)  
\\
\mathrm{s.t.}\;\;
u=\frac{\tau-\tau_i}{\Delta\tau_{\mathrm{rot}}}
\end{gathered}
\end{equation}
where $\mathrm{Exp}(\cdot)$ maps elements in the Lie algebra to the associated Lie group, and $\mathrm{Log}(\cdot)$ is its inverse operation.

\section{Methodology}
This section presents the proposed event-based continuous-time visual-inertial spatiotemporal calibration framework.

\subsection{System Overview}
\label{sect:overview}
\begin{figure}[t]
	\centering
	\includegraphics[width=\linewidth]{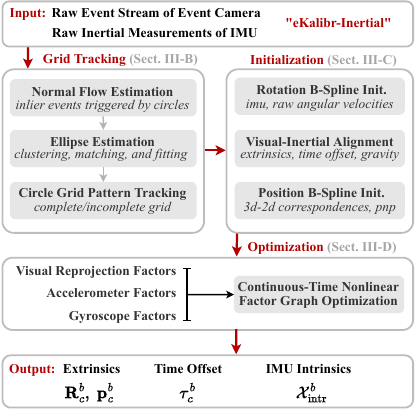}
	\caption{Illustration of the pipeline of the proposed event-based visual-inertial spatiotemporal calibration method. A detailed description of the pipeline is provided in Section \ref{sect:overview}.}
	\label{fig:overview}
	\figtabbottomspace
\end{figure}

The comprehensive framework of the proposed visual-inertial calibrator is illustrated in Fig. \ref{fig:overview}.
Given the raw asynchronous event streams from the event camera, we first perform normal flow estimation and ellipse fitting, to track both complete and incomplete circle grid patterns, see Section \ref{sect:grid_tracking}.
Subsequently, using the obtained grid patterns together with the raw inertial measurements, we recover the initial guesses of the rotation and position B-splines, extrinsics, time offset, and the gravity vector, see Section \ref{sect:initialization}.
Specifically, we first fit the rotation B-spline using raw angular velocities, and then perform visual-inertial alignment to initialize spatiotemporal parameters and the world-frame gravity vector.
Position B-spline is then recovered using camera PnP results and initialized spatiotemporal parameters.
Finally, a continuous-time-based nonlinear factor graph optimization would be carried out, with the incorporation of raw inertial measurements and extracted grid patterns from raw events, to refine all parameters to better states, see Section \ref{sect:optimization}.

The state vector of the system can be described as follows:
\begin{equation}
\mathcal{X}\triangleq\left\lbrace\mathcal{X}_{\mathrm{pos}},\mathcal{X}_{\mathrm{rot}},\rotation{c}{b},\translation{c}{b},\timeoffset{c}{b},\mathcal{X}_{\mathrm{intr}}^b,\gravity{w}\right\rbrace
\end{equation}
where $\mathcal{X}_{\mathrm{pos}}$ and $\mathcal{X}_{\mathrm{rot}}$ are translational and rotational control points defined in (\ref{equ:pos_cp}) and (\ref{equ:rot_cp}), respectively;
$\rotation{c}{b}$ and $\translation{c}{b}$ denote the extrinsic rotation and translation from $\coordframe{c}$ to $\coordframe{b}$;
$\timeoffset{c}{b}$ represents the time offset between the camera and IMU, i.e., temporal transformation $\tau^{b}=\tau^{c}+\timeoffset{c}{b}$ holds;
$\mathcal{X}_{\mathrm{intr}}^b$ denotes the IMU intrinsics defined in (\ref{equ:imu_intri});
$\gravity{w}$ represents the world-frame gravity vector, considered as a two-DoF quantity with a constant Euclidean norm.
The extrinsics and time offset are exactly the spatiotemporal parameters \emph{eKalibr-Stereo} calibrates.

\subsection{Event-Based Circle Grid Recognition and Tracking}
\label{sect:grid_tracking}
Given generated raw event streams, we first employ the event-based circle grid pattern recognition algorithm \cite{chen2025ekalibr} proposed in  \emph{eKalibr} to extract complete grid patterns for each camera.
As described in \cite{chen2025ekalibr}, we first perform event-based normal flow estimation \cite{lu2024EventBased} on the surface of active event (SAE) \cite{delbruck2008frame} and homopolarly cluster inlier events for cluster matching.
Spatiotemporal ellipses would then be estimated for each matched cluster pair for center determination of the grid circle.
Finally, temporally synchronized centers would be organized as ordered grid patterns (more details, please refer to \cite{chen2025ekalibr}).

In addition to the aforementioned grid pattern recognition algorithm \cite{chen2025ekalibr}, the incomplete grid pattern tracking module proposed in \emph{eKalibr-Stereo} \cite{chen2025ekalibr-stereo} is also employed, to improve continuity of grid tracking\footnote{
\textbf{Continuity of grid tracking}: the motion-based spatiotemporal calibration determines parameters based on the rigid-body constraint under continuous motion, thereby requiring motion state estimation from continuous tracking of the grid.
}.
Specifically, leveraging the prior knowledge of motion continuity, we construct a three-point Lagrange polynomial \cite{werner1984polynomial} for each grid circle that had been continuously tracked three times, and then predict its position in the subsequent SAE map.
When the predicted point exhibits sufficient proximity to its nearest ellipse center extracted from the subsequent SAE, we designate the newly extracted ellipse center as the corresponding position of the grid circle in the subsequent SAE.
Once enough predicted grid circles were associated with ellipse centers in the subsequent SAE map, we organized a new incomplete tracked grid pattern.
Note that, to ensure maximal tracking continuity, we would iteratively perform alternating forward and backward tracking of incomplete grid patterns until no additional ones can be tracked (more details, please refer to \cite{chen2025ekalibr-stereo}).

Note that although both the asymmetric and symmetric circle grids are supported, the asymmetric circle grid is utilized in this work, as it does not exhibit 180-degree ambiguity \cite{mathworks_calibration_patterns}.
For notational convenience, we denote all tracked grid patterns as:
\begin{equation}
\mathcal{P}\triangleq\left\lbrace
\left( \mathcal{G}_k,\tau_k\right)
\right\rbrace
\;\mathrm{s.t.}\;\;
\mathcal{G}_k\triangleq\left\lbrace
\left. \left( \bsm{x}^k_j,\bsm{p}^w_j\right) \right| 
\bsm{x}^k_j
\in\mathbb{R}^2,\bsm{p}^w_j\in\mathbb{R}^3
\right\rbrace
\end{equation}
where $\mathcal{G}_k$ denotes the $k$-th tracked grid pattern at time $\tau_k$, storing tracked 2D ellipse centers $\left\lbrace \bsm{x}^k_j\right\rbrace $ and their associated 3D grid circle centers $\left\lbrace \bsm{p}^w_j\right\rbrace $ on the board.

\subsection{Initialization}
\label{sect:initialization}
Considering the high non-linearity of continuous-time optimization, an efficient three-stage initialization procedure is designed to orderly recover initial guesses of all parameters in the estimator.

\subsubsection{Rotation B-Spline Initialization}
Given the raw body-frame angular velocity measurements from the gyroscope, the rotation B-spline could be first recovered by solving the following nonlinear least-squares problem:
\begin{equation}
\label{equ:rot_spline_recovery}
\hat{\mathcal{X}}_{\mathrm{rot}}
\gets\arg\min
\sum_{k}^{\mathcal{W}}\left\| 
\bsm{r}_{\omega}^k
\right\| ^2_{\bsm{Q}_{\omega,k}}
\end{equation}
with
\begin{equation}
\label{equ:gyro_residual}
\begin{aligned}
\bsm{r}_{\omega}^k    &\triangleq\tilde{\bsm{\omega}}_k-
\pi_\omega\left( {\bsm{\omega}}(\tau_k^b),\mathcal{X}_{\mathrm{intr}}^\omega\right) 
\\[-2pt]
{\bsm{\omega}}(\tau)&=
\left( \rotationhat{b}{b_0}(\tau)\right)^\top\cdot\angvelhat{b}{b_0}(\tau)	
\end{aligned}
\end{equation}
where $\mathcal{W}$ denotes the noisy angular velocity data sequence from the  gyroscope, in which $\tilde{\bsm{\omega}}_k$ is the $k$-th measurement at time $\tau_k^b$ stamped by the IMU clock;
$\bsm{r}_{\omega}^k$ denotes the gyroscope residual with information matrix ${\bsm{Q}_{\omega,k}}$ determined by measurement noise level;
$\pi_\omega(\cdot)$ is the gyroscope measuring function defined in (\ref{equ:imu_mearsing});
$\rotationhat{b}{b_0}(\tau)$ and $\angvelhat{b}{b_0}(\tau)$ are the ideal rotation and angular velocity at time $\tau$, analytically obtained from the rotation B-spline based on (\ref{equ:rot_bspline}), which exactly involves the rotation control points into the optimization.
Note that the gyroscope intrinsics $\mathcal{X}_{\mathrm{intr}}^\omega$ is initialized as ideal ones here, i.e., $\bsm{M}_\omega\gets\bsm{I}_{3\times 3}$ and $\bsm{b}_\omega\gets \bsm{0}_3$.
Additionally, the first IMU frame $\coordframe{b_0}$ is considered as the reference frame here.

\subsubsection{Visual-Inertial Alignment}
In this stage, the extrinsics and time offset between the camera and IMU are initialized by aligning kinematics of two sensors.
We first perform PnP for each extracted grid pattern using 2D-3D correspondences to obtain the world-frame pose sequence of the event camera.
Subsequently, the rotation-only hand-eye alignment is conducted to recover the extrinsic rotation and the time offset, which could be described as:
\begin{equation}
\label{equ:rot_hand_eye}
\left\lbrace \hat{\bsm{R}}_{c}^{b},\timeoffsethat{c}{b}\right\rbrace  \gets \arg\min\sum^{\mathcal{T}}_k
\left\| 
\bsm{r}_\mathrm{rot}^k
\right\| ^2
\end{equation}
with
\begin{equation}
\bsm{r}_\mathrm{rot}^k\triangleq
\mathrm{Log}\left( 
\rotationhat{c}{b} \cdot \rotation{c_{k\smallplus 1}}{c_{k}}
  \cdot\left( \rotation{b_{k\smallplus 1}}{b_{k}}\cdot
\rotationhat{c}{b}\right) ^\top\right) 
\end{equation}
and
\begin{equation}
\begin{aligned}
\rotation{c_{k\smallplus 1}}{c_{k}}&\triangleq
\left( \rotation{c_k}{w}\right) ^\top
\cdot\rotation{c_{k\smallplus 1}}{w}
\\[-2pt]
\rotation{b_{k\smallplus 1}}{b_{k}}&\triangleq
\left( \rotation{b}{b_0}(\tau_{k}^c+\timeoffsethat{c}{b})\right) ^\top
\cdot\rotation{b}{b_0}(\tau_{k\smallplus 1}^c+\timeoffsethat{c}{b})
\end{aligned}
\end{equation}
where $\rotation{c_k}{w}\in\mathcal{T}$ is the rotation component of the pose of the $k$-th extracted grid pattern; $\rotation{b}{b_0}(\cdot)$ is the rotation of the IMU obtained from the fitted rotation B-spline.
Note that both extrinsic rotation and time offset\footnote{
When the temporal offset between the two sensors is sufficiently small (e.g., less than 20 ms), the time delay can be directly initialized as zero and subsequently refined through the optimization in (\ref{equ:rot_hand_eye}). In contrast, when the temporal offset is relatively large, a \textbf{cross-correlation-based temporal initialization} \cite{shu2022robust} is first employed to estimate a coarse initial offset, which then serves as the input for the subsequent \textbf{rotation-only hand–eye alignment} defined in (\ref{equ:rot_hand_eye}).
Details about the cross-correlation-based temporal initialization can be found in \hyperref[sect:appendix]{Appendix}.
} can be roughly determined based on (\ref{equ:rot_hand_eye}).
Once the extrinsic rotation and time offset are recovered, the rotation B-spline can be transformed from $\coordframe{b_0}$ to $\coordframe{w}$:
\begin{equation}
\mathcal{X}_{\mathrm{rot}}\gets\rotation{b_0}{w}\cdot\mathcal{X}_{\mathrm{rot}}
\end{equation}
by computing:
\begin{equation}
\rotation{b_0}{w}\gets\rotation{c_0}{w}\cdot\left( \rotation{b}{b_0}(\tau_{0}^c+\timeoffset{c}{b})\cdot\rotation{c}{b}\right)^\top.
\end{equation}

Subsequently, the translational components of the camera poses estimated from PnP would be aligned with the integrated measurements from the accelerometer, to recover the extrinsic translation.
Through first- and second-order integration of the following kinematic constraint:
\begin{equation}
\bsm{a}(\tau)=\left(\rotation{b}{w}(\tau) \right) ^\top\cdot\left(\linacce{b}{w}(\tau)-\gravity{w}\right) 
\end{equation}
we can obtain:
\begin{equation}
\label{equ:vel_integration}
\begin{aligned}
\linvel{b}{w}(\tau+\Delta \tau)
&=\linvel{b}{w}(\tau)+
\bsm{\alpha}_{\Delta \tau}
+\gravity{w}\cdot\Delta \tau
\\
\bsm{\alpha}_{\Delta \tau}&\triangleq\int_{\tau}^{\tau+\Delta \tau}\rotation{b}{w}(t)\cdot\bsm{a}(t) \cdot \mathrm{d}t
\end{aligned}
\end{equation}
and
\begin{equation}
\label{equ:pos_integration}
\begin{aligned}
\translation{b}{w}(\tau+\Delta \tau)
&=\translation{b}{w}(\tau)+
\bsm{\beta}_{\Delta \tau}+\linvel{b}{w}(\tau)\cdot\Delta \tau+
\frac{1}{2}\cdot\gravity{w}\cdot\Delta^2\tau
\\
\bsm{\beta}_{\Delta \tau}&\triangleq\iint_{\tau}^{\tau+\Delta t}\rotation{b}{w}(t)\cdot\bsm{a}(t) \cdot \mathrm{d}t
\end{aligned}
\end{equation}
where $\linvel{b}{w}(\tau)$ and $\translation{b}{w}(\tau)$ are the linear velocity and position of the IMU at time $\tau$;
$\Delta \tau$ denotes the time distance;
$\bsm{\alpha}_{\Delta \tau}$ and $\bsm{\beta}_{\Delta \tau}$ are integration items, which can be obtained by numerical integration methods using the fitted rotation B-spline and raw accelerometer measurements.
Based on (\ref{equ:vel_integration}) and (\ref{equ:pos_integration}), the extrinsic translation and the gravity vector can be recovered simultaneously by solving the following least-squares problem:
\begin{equation}
\label{equ:pos_align}
\begin{gathered}
\left\lbrace\translationhat{c}{b},\gravityhat{w}\right\rbrace\cup\left\lbrace \linvelhat{c_{k}}{w}\right\rbrace \gets\arg\min
\sum^{\mathcal{T}}_k
\left( \left\|\bsm{r}_\mathrm{v}^k \right\| ^2+
\left\|\bsm{r}_\mathrm{p}^k\right\| ^2\right) 
\end{gathered}
\end{equation}
with
\begin{equation}
\begin{aligned}
\bsm{r}_\mathrm{v}^k&\triangleq
\linvel{b}{w}(\tau^b_{k\smallplus 1})-\linvel{b}{w}(\tau^b_k)-
{\bsm{\alpha}}_{\Delta \tau}
-\gravityhat{w}\cdot\Delta \tau
\\
\bsm{r}_\mathrm{p}^k&\triangleq
\translation{b}{w}(\tau^b_{k\smallplus 1})-\translation{b}{w}(\tau^b_k)-
{\bsm{\beta}}_{\Delta \tau}-
\linvel{b}{w}(\tau_k^b)\cdot\Delta \tau
\\&\quad-\frac{1}{2}\cdot\gravityhat{w}\cdot\left( \Delta \tau\right) ^2
\end{aligned}
\end{equation}
and
\begin{equation}
\begin{aligned}
\linvel{b}{w}(\tau_k^b)&=\linvelhat{c_k}{w}-\liehat{\angvel{b}{w}(\tau_{k}^c+\timeoffset{c}{b})}\cdot\rotation{b}{w}(\tau_{k}^c+\timeoffset{c}{b})\cdot\translationhat{c}{b}
\\
\translation{b}{w}(\tau_k^b)&=\translation{c_k}{w}-\rotation{b}{w}(\tau_{k}^c+\timeoffset{c}{b})\cdot\translationhat{c}{b}
\end{aligned}
\end{equation}
where $\translation{c_k}{w}\in\mathcal{T}$ is the camera position at time $\tau_k^c$ in $\coordframe{w}$ obtained from PnP;
Note that the world-frame camera velocities $\{\linvelhat{c_k}{w}\}$ are also estimated in (\ref{equ:pos_align}).

\subsubsection{Position B-Spline Initialization}
Finally, the (world-frame) position B-spline of the IMU can be initialized based on the estimated camera positions from PnP and recovered spatiotemporal parameters.
This can be conducted by solving the following least-squares problem:
\begin{equation}
\label{equ:pos_spline_recovery}
\hat{\mathcal{X}}_{\mathrm{pos}}
\gets\arg\min
\sum_{k}^{\mathcal{T}}\left\| 
\bsm{r}^k_{\mathrm{pos}}
\right\| ^2
\end{equation}
with
\begin{equation}
\bsm{r}^k_{\mathrm{pos}}\triangleq
\rotation{b}{w}(\tau_{k}^c+\timeoffset{c}{b})\cdot\translation{c}{b}+\translationhat{b}{w}(\tau_{k}^c+\timeoffset{c}{b})-\translation{c_k}{w}
\end{equation}
where $\translationhat{b}{w}(\tau_{k}^c+\timeoffset{c}{b})$ denotes the IMU position at $\tau_k^b$, analytically obtained from the position B-spline based on (\ref{equ:pos_bspline}), which exactly involves the position control points into the optimization.

At this stage, all spatiotemporal parameters (the extrinsics and time offset) and B-splines are initialized.
As for the intrinsics of the IMU, they are set to ideal ones directly, with identity matrices or zero vectors accordingly.

\subsection{Continuous-Time Factor Graph Optimization}
\label{sect:optimization}
Based on the extracted grid patterns and raw inertial measurements, a continuous-time batch optimization would be performed to refine all initialized parameters to better states.
Together three kinds of factors are involved in the optimization: visual reprojection factors for the camera, as well as gyroscope factors and accelerometer factors for the IMU.
\subsubsection{Visual Reprojection Factor}
Given a 2D-3D correspondence $\left( \bsm{x}^k_j,\bsm{p}^w_j\right)\in\mathcal{G}_k$, a corresponding visual reprojection residual can be constructed, which introduces the optimization of B-splines and spatiotemporal parameters.
The visual reprojection residual can be expressed as:
\begin{equation}
\bsm{r}_{c}^{k,j}\triangleq\tilde{\bsm{x}}^k_j-\pi_c\left(\bsm{p}^{c_k}_j,{\bsm{x}}^c_{\mathrm{intr}}\right)  
\end{equation}
with
\begin{equation}
\bsm{p}^{c_k}_j=\left( \transformhat{b}{w}(\tau_k^c+\timeoffsethat{c}{b})\cdot\transformhat{c}{b}\right) ^{-1}\cdot\bsm{p}^{w}_j
\end{equation}
where $\pi_c(\cdot)$ is the visual projection function defined in (\ref{equ:proj_func});
$\transformhat{b}{w}(\cdot)$ denotes the IMU pose obtained from the rotation B-spline and position B-spline.

\subsubsection{Accelerometer Factor}
Given a specific force measurement $\tilde{\bsm{a}}_k \in \mathcal{A}$, a corresponding accelerometer residual can be constructed, which introduces the optimization of B-splines, the gravity vector, and the accelerometer intrinsics.
The accelerometer residual can be expressed as:
\begin{equation}
\bsm{r}_{a}^k\triangleq
\tilde{\bsm{a}}_k-\pi_a\left(\bsm{a}(\tau_k^b),\hat{\mathcal{X}}_{\mathrm{intr}}^a \right)
\end{equation}
with
\begin{equation}
\bsm{a}(\tau)=\left(\rotationhat{b}{w}(\tau) \right) ^\top\cdot\left(\linaccehat{b}{w}(\tau)-\gravityhat{w}\right)
\end{equation}
where $\linaccehat{b}{w}(\tau)$ denotes the linear acceleration of the IMU at time $\tau$,  which can be analytically obtained from the linear scale B-spline based on (\ref{equ:pos_bspline}).

\subsubsection{Gyroscope Factor}
Given a angular velocity measurement $\tilde{\bsm{\omega}}_k \in \mathcal{W}$, a corresponding gyroscope residual can be constructed, which introduces the optimization of the rotation B-spline and gyroscope intrinsics.
The accelerometer residual has been defined in (\ref{equ:gyro_residual}).

\subsubsection{Optimization}
The final continuous-time batch optimization could be expressed as the following least-squares problem:
\begin{equation}
\begin{aligned}
\hat{\mathcal{X}}\gets
\arg\min
&\sum_{k}^{\mathcal{P}}\sum_{j}^{\mathcal{G}_k}
\rho\left(\left\|\bsm{r}_{c}^{k,j}\right\|^2_{\bsm{Q}_{c,k}}\right)
\\&+
\sum_{k}^{\mathcal{A}}\left\| \bsm{r}_{a}^k \right\| ^2_{\bsm{Q}_{a,k}}+
\sum_{k}^{\mathcal{W}}\left\| \bsm{r}_{\omega}^k \right\| ^2_{\bsm{Q}_{\omega,k}}
\end{aligned}
\end{equation}
where $\bsm{Q}_{(\cdot)}$ denotes the information matrix of the measurement;
$\rho(\cdot)$ is the Huber loss function.
The \emph{Ceres solver} \cite{Agarwal_Ceres_Solver_2022} is used for solving this nonlinear problem.
\section{Real-World Experiment}

\subsection{Equipment Setup}
\begin{figure}[t]
	\centering
	\includegraphics[width=\linewidth]{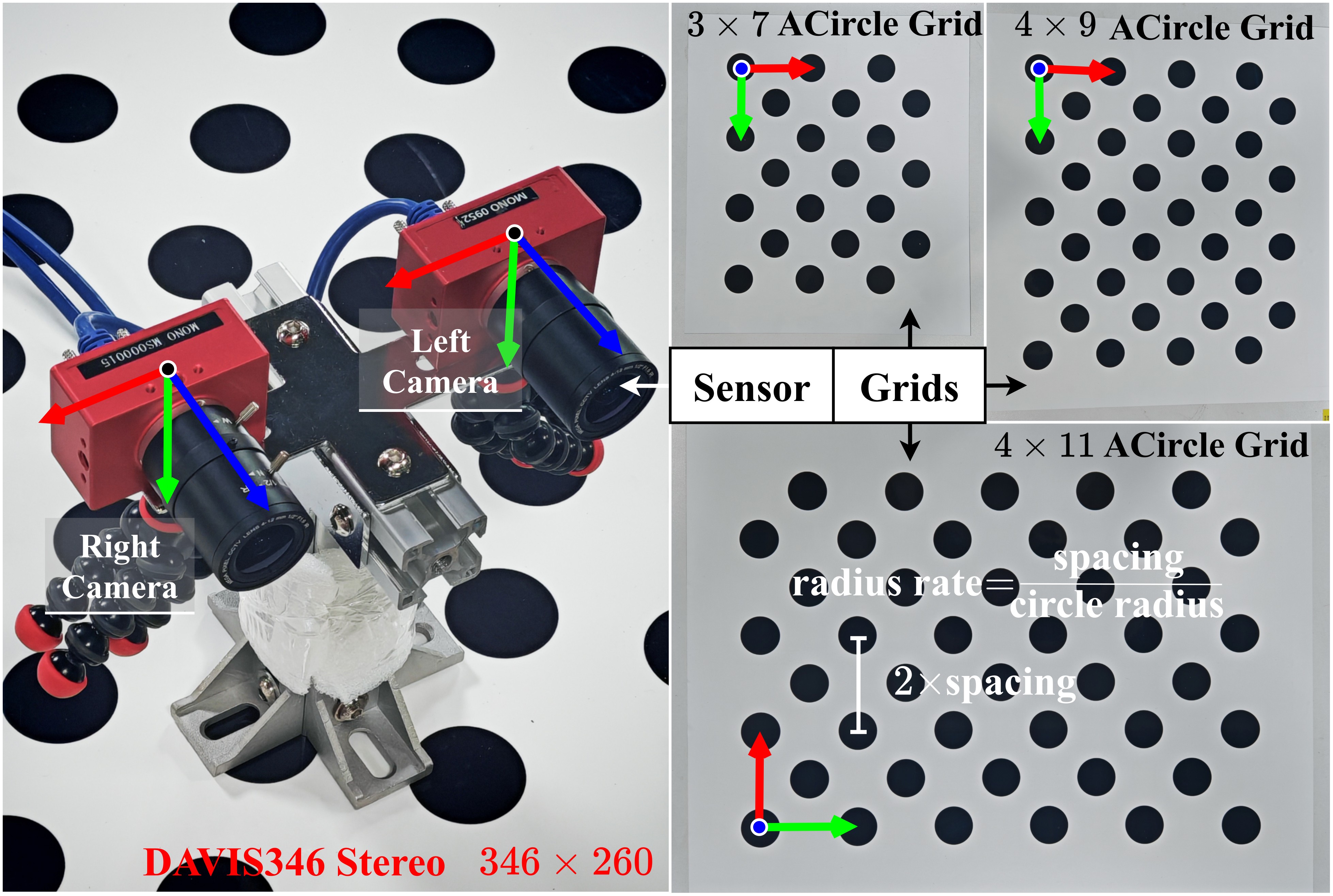}
	\caption{Stereo event camera rig (left subfigure) and three kinds of asymmetric circle grid patterns (right subfigures) utilized in real-world experiments.}
	\label{fig:setup}
\end{figure}

Fig. \ref{fig:setup} shows the self-assembled sensor suite for real-world experiments, consisting of two hardware-synchronized \emph{DAVIS346} event cameras (the resolution is 346$\times$260).
We refer to the two event cameras as the left camera (whose camera frame and built-in IMU frame denote $\coordframe{c_{\mathrm{left}}}$ and $\coordframe{b_{\mathrm{left}}}$, respectively) and the right camera (whose camera frame and built-in IMU frame denote $\coordframe{c_{\mathrm{right}}}$ and $\coordframe{b_{\mathrm{right}}}$, respectively) for convenience in subsequent description and discussion.
To ensure the comprehensiveness of the experiment, three different sizes of asymmetric circle grid patterns (3$\times$7, 4$\times$9, and 4$\times$11), as shown in Fig. \ref{fig:setup}, are used in real-world experiments.
The radius rate and spacing for all grid boards are 2.5 and 50 mm, respectively.

\subsection{Evaluation of Calibration Performance}
To comprehensively and quantitatively evaluate the spatiotemporal calibration performance of the proposed \emph{eKalibr-Inertial}, we conducted real-world Monte-Carlo experiments, where 5 sequences of 30-second data are collected for each grid board for evaluation.

Table \ref{tab:st_calib_results} summarized final spatiotemporal calibration results of the proposed \emph{eKalibr-Inertial} in real-world Monte-Carlo experiments, showing the spatiotemporal estimates and corresponding standard deviations (STDs).

\begin{table*}[]
\centering
\caption{\textbf{Spatiotemporal Calibration Results in Monte-Carlo Experiments}
\\eKalibr-Inertial could achieve high calibration accuracy and reliability
}
\label{tab:st_calib_results}
\begin{threeparttable}
\begin{tabular}{c|c|l|rrrrrr|rc}
\toprule
\multirow{3}{*}{Method}                            & \multicolumn{1}{c|}{\multirow{3}{*}{Sensor}}                        & \multicolumn{1}{c|}{\multirow{3}{*}{Board}} & \multicolumn{6}{c|}{Extrinsic}                                                                                                                                        & \multicolumn{2}{c}{Temporal}               \\ \cmidrule{4-11} 
                                                   & \multicolumn{1}{c|}{}                                               & \multicolumn{1}{c|}{}                       & \multicolumn{3}{c|}{Rotation (Euler angles, unit: degree)}                                   & \multicolumn{3}{c|}{Translation (unit: cm)}                            & \multicolumn{2}{c}{Time Offset (unit: ms)} \\ \cmidrule{4-11} 
                                                   & \multicolumn{1}{c|}{}                                               & \multicolumn{1}{c|}{}                       & \multicolumn{1}{c}{Roll} & \multicolumn{1}{c}{Pitch} & \multicolumn{1}{c|}{Yaw}              & \multicolumn{1}{c}{X} & \multicolumn{1}{c}{Y} & \multicolumn{1}{c|}{Z} & \multicolumn{1}{c}{Est.}       & Ref.      \\ \midrule\midrule
\multirow{12}{*}{\rotatebox{90}{eKalibr-Inertial}} & \multirow{4}{*}{$\underrightarrow{\mathcal{F}}_{c_\mathrm{left}}$}  &  3$\times$7                                  & 179.84$\pm$0.08          & 0.34$\pm$0.03             & \multicolumn{1}{r|}{-179.78$\pm$0.01} & -0.01$\pm$0.02        & -0.37$\pm$0.06        & -5.56$\pm$0.14         & 1.51$\pm$0.06                  & 0.00      \\
                                                   &                                                                     & 4$\times$9                                  & 179.77$\pm$0.05          & 0.42$\pm$0.04             & \multicolumn{1}{r|}{-179.91$\pm$0.04} & 0.05$\pm$0.07         & -0.22$\pm$0.15        & -5.33$\pm$0.03         & 1.23$\pm$0.18                  & 0.00      \\
                                                   &                                                                     & 4$\times$11                                 & 179.73$\pm$0.05          & 0.27$\pm$0.03             & \multicolumn{1}{r|}{-179.73$\pm$0.05} & 0.15$\pm$0.05         & -0.52$\pm$0.13        & -5.36$\pm$0.20         & 1.24$\pm$0.06                  & 0.00      \\
                                                   &                                                                     & \multicolumn{1}{c|}{Overall}                  & 179.78$\pm$0.10          & 0.34$\pm$0.07             & \multicolumn{1}{r|}{-179.81$\pm$0.06} & 0.06$\pm$0.08         & -0.37$\pm$0.17        & -5.41$\pm$0.17         & 1.33$\pm$0.17                  & 0.00      \\ \cmidrule{2-11} 
                                                   & \multirow{4}{*}{$\underrightarrow{\mathcal{F}}_{c_\mathrm{right}}$} & 3$\times$7                                  & 179.37$\pm$0.08          & -1.72$\pm$0.03            & \multicolumn{1}{r|}{179.87$\pm$0.03}  & -12.13$\pm$0.02       & -0.39$\pm$0.06        & -5.17$\pm$0.14         & 1.58$\pm$0.09                  & 0.00      \\
                                                   &                                                                     & 4$\times$9                                  & 179.35$\pm$0.05          & -1.65$\pm$0.05            & \multicolumn{1}{r|}{179.76$\pm$0.05}  & -12.06$\pm$0.09       & -0.26$\pm$0.15        & -4.93$\pm$0.11         & 1.45$\pm$0.08                  & 0.00      \\
                                                   &                                                                     & 4$\times$11                                 & 179.25$\pm$0.05          & -1.83$\pm$0.04            & \multicolumn{1}{r|}{179.95$\pm$0.04}  & -12.03$\pm$0.06       & -0.55$\pm$0.12        & -5.18$\pm$0.33         & 1.36$\pm$0.05                  & 0.00      \\
                                                   &                                                                     & \multicolumn{1}{c|}{Overall}                  & 179.32$\pm$0.10          & -1.73$\pm$0.08            & \multicolumn{1}{r|}{179.86$\pm$0.06}  & -12.07$\pm$0.08       & -0.40$\pm$0.16        & -5.09$\pm$0.25         & 1.46$\pm$0.11                  & 0.00      \\ \cmidrule{2-11} 
                                                   & \multirow{4}{*}{$\underrightarrow{\mathcal{F}}_{b_\mathrm{right}}$} & 3$\times$7                                  & 0.43$\pm$0.01            & 1.39$\pm$0.01             & \multicolumn{1}{r|}{0.24$\pm$0.01}    & -11.80$\pm$0.02       & -0.24$\pm$0.06        & 0.32$\pm$0.03          & -0.10$\pm$0.03                 & 0.00      \\
                                                   &                                                                     & 4$\times$9                                  & 0.44$\pm$0.01            & 1.39$\pm$0.01             & \multicolumn{1}{r|}{0.23$\pm$0.01}    & -11.81$\pm$0.08       & -0.14$\pm$0.05        & 0.28$\pm$0.02          & 0.11$\pm$0.12                  & 0.00      \\
                                                   &                                                                     & 4$\times$11                                 & 0.43$\pm$0.01            & 1.47$\pm$0.01             & \multicolumn{1}{r|}{0.24$\pm$0.01}    & -11.73$\pm$0.07       & -0.33$\pm$0.08        & 0.44$\pm$0.04          & -0.08$\pm$0.09                 & 0.00      \\
                                                   &                                                                     & \multicolumn{1}{c|}{Overall}                  & 0.43$\pm$0.01            & 1.41$\pm$0.03             & \multicolumn{1}{r|}{0.24$\pm$0.01}    & -11.78$\pm$0.07       & -0.24$\pm$0.10        & 0.35$\pm$0.07          & -0.02$\pm$0.13                 & 0.00      \\ \bottomrule
\end{tabular}
\begin{tablenotes} 
\item[*] All spatiotemporal parameters in this table, i.e., extrinsics and time offset, are those of the sensor with respect to the left IMU ($\coordframe{b_{\mathrm{left}}}$).
\item[*] The value in each table cell is represented as (Estimate Mean) $\pm$ (STD). A smaller STD indicates better repeatability and stability of the method.
\end{tablenotes}
\end{threeparttable}
\end{table*}


\section{Conclusion} 
In this article, we present the continuous-time-based spatiotemporal calibrator for event-based visual-inertial systems, named \emph{eKalibr-Inertial}, which is event-only and can accurately estimate both extrinsic and temporal parameters of the sensor suite.
Based on tracked grid patterns and raw inertial measurements, a three-step initialization is first performed to recover the initial guesses of all parameters in the estimator, followed by a continuous-time batch optimization to refine all parameters to the optimal states.
Extensive real-world experiments were conducted to evaluate the performance of the \emph{eKalibr-Inertial} regarding spatiotemporal calibration and computation Consumption.
The results indicate that \emph{eKalibr-Inertial} could achieve calibration accuracy comparable to frame-based visual-inertial calibrators.

\section*{Appendix}
\label{sect:appendix}
The world-frame rotations of the IMU and the camera at a given time instant are related via the extrinsic rotation:
\begin{equation}
\rotation{c}{w}(\tau)=\rotation{b}{w}(\tau)\cdot\rotation{c}{b}.
\end{equation}
By taking the first-order time derivative of the above equation, we obtain:
\begin{equation}
\angvel{c}{}(\tau)=\left( \rotation{c}{b}\right) ^\top\cdot\angvel{b}{}(\tau)
\end{equation}
where $\angvel{c}{}(\tau)$ and $\angvel{b}{}(\tau)$ denote the sensor-frame angular velocities of the camera and IMU, respectively.
Based on this fact, we have:
\begin{equation}
\Vert\angvel{c}{}(\tau)\Vert\equiv\Vert\angvel{b}{}(\tau)\Vert
\end{equation}
which implies that, \textbf{at the same time instant}, the angular velocity norms of the two rigidly connected sensors are expected to be identical.
This allows us to recover the time offset initials using the cross correlation technique.
Specially, given two angular velocity sets of two sensors, denoted as $\mathcal{W}_b\triangleq\left\lbrace\angvel{b_i}{} \right\rbrace $ and $\mathcal{W}_c\triangleq\left\lbrace\angvel{c_k}{} \right\rbrace $, the time offset step $s$ can be estimated be solving the following optimization problem:
\begin{equation}
\hat{s}\gets
\arg\max
\sum^{\mathcal{W}_b,\mathcal{W}_c}_{i,k}
\Vert\angvel{c_k}{}\Vert\cdot
\Vert\angvel{b_{i^*}}{}\Vert
\end{equation}
with
\begin{equation}
\angvel{b_{i^*}}{}\;\mathrm{has}
\;i^*\gets\arg\min\vert \hat{\tau}^b_i-\left( \tau^c_k+\hat{s}\cdot\Delta\tau^b_{\mathrm{avg}}\right) \vert
\end{equation}
where $\Delta\tau^b_{\mathrm{avg}}$ denotes the average time distance between two consecutive inertial measurements.
Finally, the time offset can be obtained: $\timeoffset{c}{b}\gets s\cdot\Delta\tau^b_{\mathrm{avg}}$.

While the IMU directly measures body-frame angular velocities, the camera only provides bearing measurements.
In order to recover the angular velocities of the camera, the discrete camera rotations are first expressed as a continuous-time rotation function, and the angular velocities are then computed by taking its first-order time derivative.
For example, the three-point Lagrange Polynomial for $\mathrm{SO(3)}$ can be expressed as:
\begin{equation}
\begin{aligned}
	L_3(\tau) &= \bsm{R}_0 \oplus
			a(\tau)\cdot\left( \bsm{R}_1 \ominus \bsm{R}_0\right)  \oplus
			b(\tau)\cdot\left( \bsm{R}_2 \ominus \bsm{R}_1\right)
		\\&=\bsm{R}_0\cdot\Exp{a(\tau)\cdot\Log{\bsm{R}_0^\top\cdot\bsm{R}_1}}
		\\&\quad\;\cdot\Exp{b(\tau)\cdot\Log{\bsm{R}_1^\top\cdot \bsm{R}_2}}
	\end{aligned}
\end{equation}
with
\begin{equation}
	\begin{aligned}
	a(\tau)&\triangleq\frac{(\tau - \tau_0)((\tau_2 - \tau_0)-(\tau-\tau_1))}{(\tau_0 - \tau_2)(\tau_0 - \tau_1)}
	\\
	b(\tau)&\triangleq\frac{(\tau - \tau_0)(\tau - \tau_1)}{(\tau_2 - \tau_0)(\tau_2 - \tau_1)}
	\end{aligned}
\end{equation}
where $(\tau_0,\rotation{0}{})$,$(\tau_1,\rotation{1}{})$, and $(\tau_2,\rotation{2}{})$ are three consecutive rotations of the camera.

\section*{CRediT Authorship Contribution Statement}
\label{sect:author_contribution}
\textbf{Shuolong Chen}: Conceptualisation, Methodology, Software, Validation, Original Draft.
\textbf{Xingxing Li}: Supervision.
\textbf{Liu Yuan}: Data Curation, Review and Editing.
	
\bibliographystyle{IEEEtran}
\bibliography{reference}

\end{document}